\RequirePackage{silence}
\documentclass{ceurart}

\ExplSyntaxOn
\AddToHook{env/Abstract/end}{\box_set_wd:Nn \g_ceur_key_box { 0pt }}
\ExplSyntaxOff
\makeatletter
\def\@copyrightLine{%
  \begingroup
  \hbadness=10000
  \begin{minipage}[c]{\ccLogoWidth}
    \doclicenseImage[imagewidth=\ccLogoWidth]
  \end{minipage}
  \nobreak\enskip
  \begin{minipage}[c]{\linewidth - \ccLogoWidth - 2\footnotemargin - .5em}
    \tiny
    \textcopyright\ \@copyrightyear\ %
    \ifx\@copyrightclause\@empty%
      \doclicenseText
    \else\@copyrightclause%
    \fi
  \end{minipage}
  \endgroup
}
\makeatother

\usepackage{listings}
\usepackage{float}
\usepackage{placeins}
\begin{document}

\copyrightyear{2026}
\copyrightclause{Copyright for this paper by its authors.
  Use permitted under Creative Commons License Attribution 4.0
  International (CC BY 4.0).}

\conference{ExplAI-2026: 2nd Workshop on Advanced AI in Explainability and
  Ethics for the Sustainable Development Goals, November 13, 2026,
  Khmelnytskyi, Ukraine}

\title{Do Language Models Feel the Sentiment? Investigating Emotion Vectors}

\author[1]{Ali Hassan Khan}[%
orcid=0009-0001-2718-4431,
]
\fnmark[1]

\author[1]{Saad Bin Ather}[%
orcid=0009-0009-3555-5528,
email=l230838@lhr.nu.edu.pk,
]
\cormark[1]
\fnmark[1]

\author[1]{Muhammad Saif}
\fnmark[1]

\author[1]{Maryam Bashir}[%
orcid=0000-0003-3684-7185,
]

\author[1]{Hajra Waheed}[%
orcid=0000-0003-0168-0063,
]

\address[1]{National University of Computer
and Emerging Sciences, Lahore, Pakistan}

\cortext[1]{Corresponding author.}
\fntext[1]{These authors contributed equally.}

\begin{abstract}
Emotion vectors are directions in a language model's
internal representations that are associated with
emotional concepts. Prior work shows that these
vectors can influence model behavior, but their
role in sentiment classification remains unclear.
This study investigates emotion directions in
Qwen2.5-0.5B-Instruct and Qwen2.5-3B-Instruct.
Eight directions extracted from GoEmotions are
evaluated through sentiment probing, activation
interventions, and supervised correction.
Emotion-only probes achieve 88.66\% and 90.70\%
accuracy on binary TweetEval. In exploratory
matched-ablation experiments, raw emotion-space
removal causes 2.39 and 1.10 percentage points
more accuracy loss than matched isotropic
random-subspace removal
for 0.5B and 3B, respectively. All four TweetEval
raw and denoised effects are significant after
Holm adjustment across eight comparisons.
An exploratory SST-2 extension shows smaller
raw excess losses of 0.66 and 0.57 points,
with no effects surviving Holm adjustment.
Frozen sentiment-correction
classifiers are also tested on 20,000 Amazon
Polarity reviews. Emotion-based correction
improves accuracy over uncorrected Qwen, but
does not consistently outperform margin-only
baselines. These results distinguish sentiment
information available in model representations,
the effects of changing those representations,
and their practical value for classification.
They do not establish a uniquely emotional
mechanism.
\end{abstract}

\begin{keywords}
Emotion Vectors \sep
Mechanistic Interpretability \sep
Sentiment Analysis \sep
Representation Geometry \sep
Causal Interventions \sep
Cross-Domain Generalization
\end{keywords}

\maketitle

\section{Introduction}
\label{sec:introduction}

Emotion recognition and sentiment classification
concern related but distinct properties of text.
A language model may encode information associated
with an emotion without using it to make a sentiment
prediction. Likewise, an accurate supervised probe
does not establish how the original model reaches its
predictions. Emotion representations therefore need
to be examined in terms of accessible information,
intervention effects, and predictive utility. This
distinction matters for trustworthy AI because it
limits what representation-based explanations can
claim about model decisions.

Sofroniew et al. \cite{sofroniew2026emotion}
showed that emotion-related representations in
Claude Sonnet 4.5 can influence model behavior.
Sentiment-interpretability research has also
identified polarity directions with causal effects
\cite{tigges2024sentiment}. The present study
examines how emotion directions derived from
an independent emotion dataset relate to sentiment
predictions in two Qwen2.5 models. Geometric
analysis, supervised probes, steering, and
subspace ablation address different aspects of
this relationship. Perturbation-matched random
controls test whether ablation effects exceed
those of comparably sized activation changes.
Supervised correction and frozen transfer then
test the practical value of emotion projections.
Throughout, emotion vectors refer to directions
associated with labeled emotional content, not
subjective emotional experience.

The main findings and contributions are:

\begin{enumerate}
    \item Emotion-derived features recover substantial
    sentiment information, although full-state probes
    perform better. Neutral-PCA denoising improves
    emotion validation and valence alignment, but
    does not consistently improve sentiment probing.

\item Steering and subspace removal change
sentiment predictions. Under matched
perturbations, all four TweetEval raw and
denoised effects survive Holm correction
across eight comparisons. None of the four
exploratory SST-2 effects survives correction.

\item Emotion features improve some trained
sentiment-correction classifiers. On Amazon
Polarity, the frozen correction heads improve
over uncorrected Qwen, but do not consistently
outperform margin-only correction. 
Raw valence steering remains below the
maximum random-direction response, whereas
denoised valence exceeds it.

\end{enumerate}

\section{Related Work}
\label{sec:related_work}

Mechanistic interpretability studies have
identified emotion-related representations
in language models. Sofroniew et al.
\cite{sofroniew2026emotion} identified emotion
vectors in Claude Sonnet 4.5 and showed
effects on preferences, sycophancy, and
reward hacking. Reichman et al.
\cite{reichman2026emotions} identified a
low-dimensional emotion space across contexts
and languages using emotionally rewritten
sentences. Maheswaran and Desarkar
\cite{maheswaran2026unified} examined
cross-dataset and cross-prompt transfer and
reported limits in emotion reasoning.
Sun et al. \cite{sun2026valence} studied
valence and arousal representations and
their effects on the emotional tone of
generated text. These studies show
emotion-related structure and behavioral effects.

Other studies focus on sentiment representations.
Tigges et al. \cite{tigges2024sentiment}
identified a linear sentiment direction
that affects model predictions.
Di Palma et al. \cite{dipalma2025feelings}
studied layer-dependent sentiment information
in LLaMA models. They showed that probe
performance depends on layer selection and
representation pooling. They also separated
recoverable information from model predictions.
Du et al. \cite{du2026usage} examined
context-dependent sentiment representations,
including effects of tone and audience.

Radwan \cite{radwan2026nine} investigated an
emotion-centroid valence direction and its
relationship with sentiment classification.
The study compared direction removal with
random-direction controls. Our work extends
this comparison to eight independently
extracted emotion directions and their shared
subspace. It uses rank-matched and
perturbation-energy-matched controls.

Probing requires careful interpretation.
Hewitt and Liang \cite{hewitt2019designing}
used control tasks to separate encoded
information from probe capacity.
Elazar et al. \cite{elazar2021amnesic}
examined whether removing recoverable
information changes model behavior.
Related methods include iterative nullspace
projection \cite{ravfogel2020null} and
linear concept erasure
\cite{belrose2023leace}. These studies
support the use of behavioral interventions
alongside supervised probes. Successful
decoding alone does not establish that
the original model uses the information.

Activation interventions provide further
evidence. Rimsky et al.
\cite{rimsky2024steering} used contrastive
activation addition to steer model behavior.
Tak et al. \cite{tak2025emotion} studied
emotion inference representations and
their causal effects. Wang et al.
\cite{wang2026circuits} examined neurons
and attention heads linked to emotional
expression. Fang et al.
\cite{fang2026deeper} combined probing
with interventions and reported stronger
effects from targeted changes than random
changes. These studies mainly address
emotion recognition, inference, or expression,
rather than sentiment classification.

Equal-rank subspace removal does not
necessarily produce equal activation changes.
Perturbation-energy matching provides a
more controlled comparison but cannot
establish a uniquely emotional mechanism.
Effects may also reflect sentiment
information shared with other directions.

Our study combines emotion-space geometry,
sentiment probing, steering, and subspace
ablation to separate accessible information
from intervention effects. Controls include
rank-matched and energy-matched removal.
Supervised sentiment correction and frozen
Amazon transfer test predictive utility
without assuming a uniquely emotional
mechanism.

\section{Methodology}
\label{sec:methodology}
\label{subsec:method_overview}

\begin{figure}[!htbp]
\centering
\includegraphics[width=\linewidth]{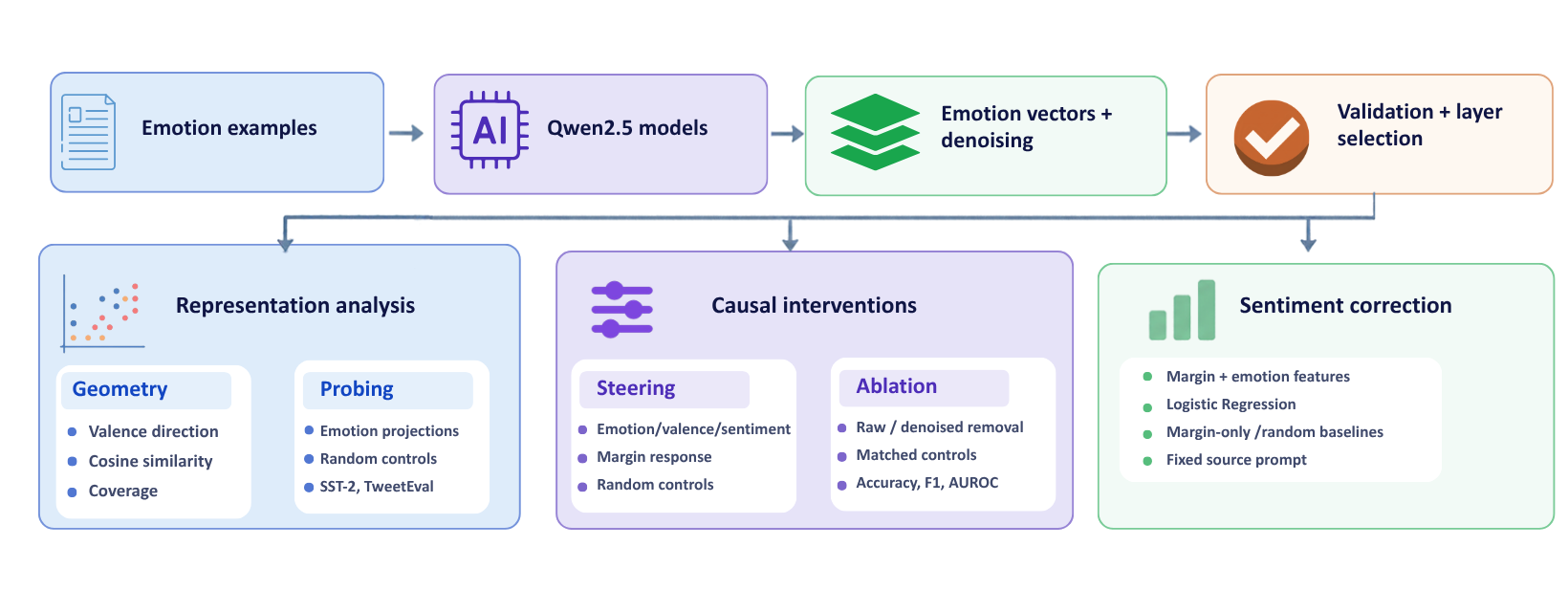}
\caption{Overview of the study. Emotion directions from
GoEmotions are used for representation analysis,
interventions, and sentiment correction. The correction
classifiers are also tested on Amazon without retraining.}
\label{fig:method_pipeline}
\end{figure}

This study examines emotion directions in frozen
language models. Eight directions are extracted
from GoEmotions, and a layer is selected for each
model through emotion validation. The directions
are evaluated using geometry, sentiment probes,
and activation interventions. Finally, supervised
correction tests whether emotion projections add
information beyond the model's output sentiment
margin. Random-direction controls provide
comparisons throughout. Figure~\ref{fig:method_pipeline}
summarizes the workflow.

\subsection{Emotion Directions and Layer Selection}
\label{subsec:method_extraction}

The experiments use frozen Qwen2.5-0.5B-Instruct
and Qwen2.5-3B-Instruct \cite{qwen2025technical}.
Emotion directions are extracted from 7,417 strict
single-label examples in the official GoEmotions
training split \cite{demszky2020goemotions}.
The eight emotions are joy, love, excitement,
gratitude, sadness, anger, fear, and disgust.

For each text, nonspecial and nonpadding raw-text
tokens are mean-pooled from the residual stream
after each decoder block. This gives one vector,
$\mathbf h_i^{(l)}$, for example $i$ at layer $l$.
For each emotion, its mean vector is compared with
the unweighted mean of the other seven emotion
class means:

\begin{equation}
\boldsymbol{\mu}_{e}^{(l)}
=\frac{1}{N_e}\sum_{i:y_i=e}\mathbf h_i^{(l)},
\qquad
\mathbf v_e^{(l)}
=\boldsymbol{\mu}_e^{(l)}
-\frac{1}{|\mathcal E|-1}
\sum_{\substack{k\in\mathcal E\\k\ne e}}
\boldsymbol{\mu}_k^{(l)}.
\label{eq:emotion_vector}
\end{equation}

Here, $N_e$ is the number of examples labeled
with emotion $e$. The set $\mathcal E$ contains
the eight emotions. Thus, $\mathbf v_e^{(l)}$
points from the average of the other emotions
toward emotion $e$.

The directions form the matrix
$\mathbf V^{(l)}=[\mathbf v_1^{(l)},\ldots,\mathbf v_8^{(l)}]$.
The unnormalized contrasts sum to zero, so their
span has rank at most seven. Subspace ablation uses
an SVD-derived orthonormal basis, with a relative
singular-value tolerance of $10^{-5}$.

Neutral-PCA denoising tests whether removing
patterns common in neutral text changes the
emotion directions. At each layer, PCA is fitted
to 12,810 separate neutral examples from the
official GoEmotions training split. These examples
are filtered to exclude normalized-text overlap
with the emotion extraction and validation pools.
The smallest number of leading components explaining
at least 50\% of the neutral activation variance
is retained. Their projection is then removed
from each emotion contrast:

\begin{equation}
\widetilde{\mathbf v}_e^{(l)}
=
\left(
\mathbf I-\mathbf U_k^{(l)}
\mathbf U_k^{(l)\top}
\right)\mathbf v_e^{(l)}.
\label{eq:denoising}
\end{equation}

Here, $\mathbf I$ is the identity matrix.
$\mathbf U_k^{(l)}$ contains the $k$ orthonormal
PCA components. The product
$\mathbf U_k^{(l)}
\mathbf U_k^{(l)\top}\mathbf v_e^{(l)}$
is the part removed from the emotion direction.

Raw and denoised directions are kept as separate
conditions. Denoising may reduce shared neutral-text
variation, but it may also discard sentiment-relevant
information. Their geometry, probing, steering,
ablation, and correction results are therefore
compared rather than assuming denoising is beneficial.

Emotion discrimination is measured by AUROC on
held-out GoEmotions examples and by transfer
evaluation on EmpatheticDialogues
\cite{rashkin2019towards}. Both raw and denoised
directions are evaluated; only denoised directions
are used to select layers. Selection uses the
seven emotions shared across these datasets,
excluding love.

For each eligible center layer $l$, the AUROC
for each emotion is first averaged over layers
$l-1$, $l$, and $l+1$. The mean ($A$) and 25th
percentile ($Q$) are then calculated across the
seven smoothed emotion AUROCs for each dataset.
The selection score is

\begin{equation}
S^{(l)}
=
0.7\min\{A_{\mathrm{GO}}^{(l)},A_{\mathrm{ED}}^{(l)}\}
+
0.3\min\{Q_{\mathrm{GO}}^{(l)},Q_{\mathrm{ED}}^{(l)}\}.
\label{eq:layer_selection_score}
\end{equation}

Here, GO and ED denote GoEmotions and
EmpatheticDialogues. Taking the lower
score across datasets favors layers that
perform consistently on both, while the
25th-percentile term accounts for
weaker-performing emotions.

\begin{figure}[!htbp]
\centering
\includegraphics[width=\linewidth]{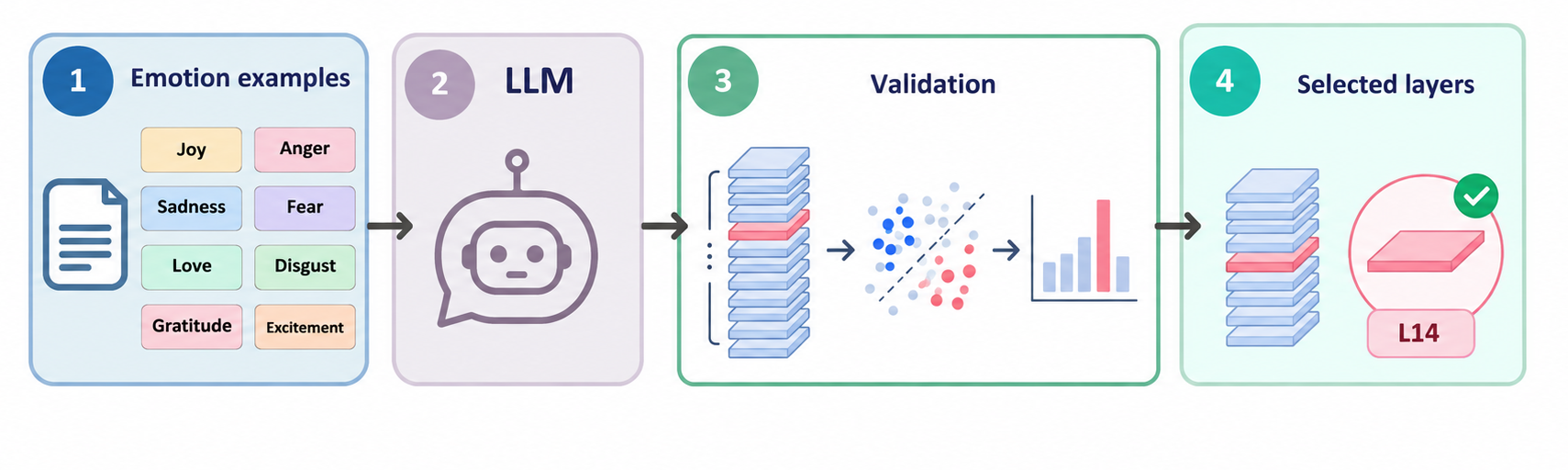}
\caption{Layer selection using emotion validation.
Layers 14 and 18 are selected for Qwen2.5-0.5B
and Qwen2.5-3B, respectively.}
\label{fig:layer_selection}
\end{figure}

The highest-scoring window determines its center
layer. Boundary layers cannot be selected as centers;
ties favor the earlier layer. This procedure selects
layer 14 for Qwen2.5-0.5B and layer 18 for
Qwen2.5-3B. The layers remain fixed in all later
experiments. Downstream sentiment labels and
accuracy are not used for layer selection.
Figure~\ref{fig:layer_selection} illustrates
the process.

\FloatBarrier

\subsection{Emotion--Sentiment Geometry and Probing}
\label{subsec:method_relationship}

Emotion--sentiment geometry and probing test whether
these directions contain positive--negative sentiment
information. All analyses use the frozen model and
the layer selected by emotion validation.

Joy, love, excitement, and gratitude form the
positive emotion group, $\mathcal E^+$; sadness,
anger, fear, and disgust form the negative group,
$\mathcal E^-$. The difference between the two
group means defines an emotion-based valence
direction:

\begin{equation}
\mathbf v_{\mathrm{val}}^{(l)}
=
\frac14\sum_{e\in\mathcal E^+}\mathbf v_e^{(l)}
-
\frac14\sum_{e\in\mathcal E^-}\mathbf v_e^{(l)}.
\label{eq:valence}
\end{equation}

Here, $l$ is the selected layer. Separate valence
directions are calculated from raw and denoised
emotion contrasts. The groups are defined without
using downstream sentiment labels.

A reference sentiment direction is derived from
sentiment-training examples passed through the
same frozen model. The difference between the
mean positive and negative pooled representations
at the selected layer defines

\begin{equation}
\mathbf d_{\mathrm{sent}}^{(l)}
=
\boldsymbol\mu_{\mathrm{pos}}^{(l)}
-
\boldsymbol\mu_{\mathrm{neg}}^{(l)}.
\label{eq:sentiment_direction}
\end{equation}

Here, $\boldsymbol\mu_{\mathrm{pos}}^{(l)}$
and $\boldsymbol\mu_{\mathrm{neg}}^{(l)}$
are the mean representations of the positive
and negative training examples. This direction
uses training labels, but no evaluation labels.
It is a reference for comparison with the
emotion directions. It is different from the
model's output sentiment margin.

Two geometric comparisons are made. Cosine
similarity measures alignment between the
emotion-based valence and sentiment directions.
Coverage measures the fraction of the sentiment
direction within the emotion subspace:

\begin{equation}
\operatorname{Coverage}
=
\frac{
\|\mathbf Q_E\mathbf Q_E^\top
\mathbf d_{\mathrm{sent}}\|_2^2
}{
\|\mathbf d_{\mathrm{sent}}\|_2^2
}.
\label{eq:coverage}
\end{equation}

Here, $\mathbf Q_E$ is an orthonormal basis for
the raw or denoised emotion space. The numerator
measures the squared norm of the sentiment
direction projected into this space; the denominator
is its total squared norm. Coverage ranges from
0 to 1 and is also compared with random
rank-matched spaces.

Sentiment probes test whether the emotion scores
of individual texts predict polarity. Each text
is encoded by the frozen model, and its mean-pooled
selected-layer representation is projected onto
the eight emotion directions. This produces eight
scores per example.

L2-regularized logistic-regression probes are
trained on raw and denoised scores separately.
Comparison probes use random directions or the
full pooled hidden state. All probes are trained
with sentiment labels but exclude Qwen's output
margin. Training-only feature standardization is
used. The regularization parameter is selected
on validation macro-$F_1$ from
$C\in\{0.001,0.01,0.1,1,10,100,1000\}$.
These tests measure recoverable sentiment
information, not whether unmodified Qwen uses it.

\subsection{Steering and Subspace Ablation}
\label{subsec:method_steering}

Two interventions test the effect of emotion
directions on model outputs. Steering adds a
direction to activations, while ablation removes
an activation component within the emotion
subspace. Both use the fixed selected layer
of each frozen Qwen model.

At each valid raw-text token, excluding prompt,
special, and padding tokens, steering adds or
subtracts a normalized direction
\cite{rimsky2024steering}:

\begin{equation}
\mathbf h_t'
=
\mathbf h_t
+
s\eta\|\mathbf h_t\|_2
\frac{\mathbf v}{\|\mathbf v\|_2},
\qquad
s\in\{-1,+1\},
\quad
\eta\in\{0.01,0.025,0.05\}.
\label{eq:steering}
\end{equation}

Here, $\mathbf h_t$ is the original token
activation, $\mathbf h_t'$ is the changed
activation, and $\mathbf v$ is the direction
being tested. The sign $s$ selects addition
or subtraction. The three values of $\eta$
make the change equal to 1\%, 2.5\%, or 5\%
of the original activation norm. In the
implementation, $\eta=0.1\alpha$, where
$\alpha\in\{0.1,0.25,0.5\}$.

Steering tests raw and denoised emotion directions,
emotion-based valence, and the sentiment reference
from Section~\ref{subsec:method_relationship}.
Fifty independently sampled Gaussian-QR random
directions provide controls. Each direction is
unit normalized before the same steering strengths
are applied. These tests match intervention size,
not the resulting effect on model outputs.

The sentiment margin, $M(x)$, is the log-probability
of the positive first-answer token minus that of
the negative first-answer token. Each candidate
label is required to tokenize as one token. The
two-sided steering response is half the difference
between the margins after positive and negative
steering. This margin is a relative label score,
not calibrated confidence or an accuracy change.

\label{subsec:method_ablation}
Subspace ablation removes an activation component
in the emotion space
\cite{ravfogel2020null,belrose2023leace}.
Let $\mathbf Q_E$ denote
its orthonormal basis. The removal fraction
$\beta$ defines the intervention:

\begin{equation}
\mathbf h_t'
=
\mathbf h_t
-
\beta\mathbf Q_E\mathbf Q_E^\top\mathbf h_t,
\qquad 0\leq\beta\leq1.
\label{eq:ablation}
\end{equation}

Full removal uses $\beta=1$. Raw and denoised
emotion spaces are tested separately against
random rank-seven spaces. Each paired comparison
uses the same examples and candidate-label
prediction rule. Ablation performance is primarily reported using
classification accuracy. Matched comparisons additionally
report excess accuracy loss relative to random-subspace
removal, with paired 95\% confidence intervals.

TweetEval intervention evaluations use
positive and negative ground-truth examples,
excluding neutral examples. The original
two-label instruction specifies positive
and negative sentiment. Binary predictions
are determined by comparing the two
candidate-token log-probabilities.
Accuracy and macro-$F_1$ are computed
over these two sentiment classes.

\label{subsec:method_distortion}
Removing subspaces of equal rank can change
activations by different amounts. A second
experiment therefore matches perturbation energy.
For each valid text token, the squared norm
of the activation change is divided by the
squared norm of its original activation. The
mean over valid tokens is the example's
normalized perturbation energy.

Full random-space removal sets an energy target.
For paired emotion-space removal, one $\beta$
is chosen per example through 16 bisection steps
to match the measured target. The same $\beta$
applies to every valid text token in that example.
Matching uses no sentiment labels. If full emotion
removal cannot reach the target, $\beta=1$ is
used and the mismatch is recorded.

Excess accuracy loss is the performance loss
after matched emotion-space removal minus the
loss after random-space removal. Fifty paired
random rank-seven controls are used per model.
Confidence intervals are obtained from paired
example-and-seed bootstrap resampling. Energy
matching controls the size of activation changes,
but not their direction or downstream effects.
The historical TweetEval analysis uses 600
previously inspected examples and is exploratory.
An SST-2 extension uses a separate 600-example
balanced internal holdout from official training;
its analysis is also treated as exploratory.

\subsection{Sentiment Correction}
\label{subsec:method_correction}

The sentiment probes measure accessible information.
Sentiment correction tests whether emotion features
improve predictions beyond the output margin.
Separate L2-regularized logistic-regression
classifiers are fitted for each Qwen model and
source dataset. Each source setting uses 14,186
training examples: 7,093 positive and 7,093
negative. The Qwen weights and emotion directions
remain frozen.

SST-2 contains sentences from movie reviews
\cite{socher2013sentiment,wang2018glue}, while
TweetEval contains tweets
\cite{barbieri2020tweeteval,rosenthal2017semeval}.
Binary sentiment experiments on both datasets evaluate positive and negative ground-truth examples.

Four prompt identifiers, P1–P4, are used for prompt-sensitivity analysis. The templates below show their standard forms. In each template, the
\texttt{\{labels\}} placeholder denotes positive or negative for both SST-2 and TweetEval. The
\texttt{\{text\}} placeholder denotes the input example. A blank line separates the instructions from the
\texttt{Text:} line.

\begin{center}
\footnotesize
\begin{tabular}{@{}p{0.47\linewidth}@{\hspace{0.03\linewidth}}p{0.47\linewidth}@{}}

\textbf{P1}\par
\texttt{Classify the sentiment of this text as
\{labels\}.}\par
\texttt{Reply with only one label:
\{labels\}.}\par
\smallskip
\texttt{Text: \{text\}}

&
\textbf{P2}\par
\texttt{Determine whether the following text
expresses \{labels\} sentiment.}\par
\texttt{Answer only: \{labels\}.}\par
\smallskip
\texttt{Text: \{text\}}

\\[8pt]

\textbf{P3}\par
\texttt{Read the text and identify its
sentiment.}\par
\texttt{Output exactly one word:
\{labels\}.}\par
\smallskip
\texttt{Text: \{text\}}

&
\textbf{P4}\par
\texttt{Is the sentiment in this text
\{labels\}?}\par
\texttt{Return only the label, with no
explanation.}\par
\smallskip
\texttt{Text: \{text\}}

\end{tabular}
\end{center}

All sentiment-correction and Amazon transfer experiments use their saved source-specific P1 instructions. Candidate-token log-probabilities provide the positive-minus-negative sentiment margin $M(x)$. The correction classifiers produce binary sentiment predictions. These experiments are distinct from prompt-sensitivity evaluation and intervention-based predictions.

Each correction classifier receives the margin
and eight emotion scores from the selected layer.
For text $x$, the token activations are mean-pooled
to obtain $\overline{\mathbf h}(x)$. The score
for emotion $e$ is

\begin{equation}
z_e(x)
=
\frac{
\overline{\mathbf h}(x)^\top\mathbf v_e
}{
\|\mathbf v_e\|_2
}.
\label{eq:emotion_score}
\end{equation}

This score measures the representation's
projection onto emotion direction $\mathbf v_e$.
The feature vector is
$[M(x),z_1(x),\ldots,z_8(x)]$.
Raw and denoised directions are evaluated
with separate classifiers.

Comparison heads use the margin alone or the
margin with 50 saved random rank-seven feature
controls. The eight mean-contrast emotion
directions sum to zero before normalization,
so their subspace has rank at most seven.
The random controls therefore match the
emotion-space rank rather than the number
of projection scores.

Feature transformations and standardization
are fitted on source training data only.
The logistic-regression parameter $C$ is
selected using validation balanced accuracy,
followed by macro-$F_1$. Ties are resolved
in favor of the smaller $C$. Test labels
are not used to fit the final classifiers
or set an evaluation-specific decision
threshold.

\subsection{Cross-Domain Transfer and Statistical Analysis}
\label{subsec:method_transfer}
\label{subsec:method_statistics}

Frozen cross-domain transfer tests whether
source-trained correction heads work on a
different dataset. A fixed, label-balanced sample
of 20,000 reviews (10,000 per class) is drawn
from the official Amazon Polarity test set
\cite{zhang2015character}. Target labels determine
the sample's class balance but are not used
to fit classifiers or choose hyperparameters.

SST-2- and TweetEval-trained classifiers are
applied for each Qwen model, giving four source--model
settings. Every setting retains its P1 prompt,
model weights, layer, feature transforms, classifier
weights, and source-trained decision rule. Amazon
labels are joined only after predictions are frozen.
Both SST-2 and TweetEval use positive and negative
labels in their prompts and all correction classifiers
produce binary sentiment predictions.

For each setting, uncorrected Qwen is compared
with margin-only, raw-emotion, denoised-emotion,
and random-feature correction. Random-control
results average 50 source-trained classifiers.
Accuracy, macro-$F_1$, balanced accuracy, and AUROC are computed, with accuracy presented as the primary cross-domain evaluation metric.

Evaluation uncertainty is estimated by bootstrap
resampling. Compared methods share paired example
resamples, with class balance preserved. The
Amazon evaluation uses 2,000 fixed-seed resamples.
Random-seed variation is reported separately
from example-resampling uncertainty. The matched
ablation analysis also resamples paired random
seeds when estimating its confidence intervals.

For matched ablation, 50,000 class-stratified
paired example-and-seed bootstrap resamples
are used on the fixed evaluation predictions.
Approximate two-sided null-centered bootstrap
$p$-values are computed with a plus-one correction
and adjusted using Holm--Bonferroni across eight
comparisons. Nominal 95\% percentile intervals
are reported for individual comparisons.
Bonferroni basic-bootstrap intervals with
99.375\% marginal coverage target approximate
95\% family-wise coverage across the eight tests.
These are not Holm-adjusted intervals, and their
coverage depends on the bootstrap approximation.

Source-domain tests and previously inspected
intervention samples are treated as exploratory.
Amazon confidence intervals remain nominal;
its predictions were frozen before target-label
outcome evaluation, although labels were used
to construct a balanced sample.

The SST-2 and TweetEval classifiers
are trained on different source datasets,
but both use positive and negative
sentiment labels. Their cross-domain
results are reported separately for
each source-trained setting.

\subsection{Reproducibility and Availability}
\label{subsec:reproducibility}

Later-stage records identify Qwen2.5-0.5B-Instruct
and Qwen2.5-3B-Instruct revisions
\texttt{7ae557604adf67be50417f59c2c2f167def9a775}
and \texttt{aa8e72537993ba99e69dfaafa59ed015b17504d1},
respectively. Correction and transfer explicitly
pin these revisions.

Recorded software includes Python~3.12.4,
PyTorch~2.14.0+cu126, Transformers~4.57.6,
NumPy~2.5.3, scikit-learn~1.9.1, and
NNsight~0.5.15. Experiments used an NVIDIA
RTX~A4000 (16,376 MiB VRAM). The recorded
driver was 596.71; PyTorch used a cu126 build.

Both tokenizers encode the exact lowercase,
unspaced strings \texttt{positive},
\texttt{negative}, and \texttt{neutral}
as single tokens with IDs 30487, 42224,
and 59568, respectively. Leading-space
variants have different IDs. Emotion extraction
used raw text, while sentiment experiments
applied Qwen chat templates with generation
prefixes. 
No extraction seed was recorded. Sampling used
seed~42, steering used~43, random controls used
1000--1049 or 2000--2049, and matched-ablation
bootstrap used seeds~45/46 with 50,000 resamples.
Correction and Amazon bootstrap used seed~47.
Recorded Amazon batch computation totaled
approximately 51.7 minutes, excluding overhead.

Source code is available at
\url{https://github.com/Alihassan0155/Do-Language-Models-Feel-the-Sentiment}.
Both 600-example identifier lists are retained
locally but are not publicly released. Datasets,
predictions, emotion vectors, trained heads,
and experimental outputs are also excluded.

\section{Experiments and Results}
\label{sec:experiments}

We evaluate frozen Qwen2.5-0.5B-Instruct and
Qwen2.5-3B-Instruct, selecting layers 14 and 18
using GoEmotions validation. Prompt sensitivity
uses 12,000 SST-2 training and 872 validation
examples, alongside 6,347 binary TweetEval
test examples. Generated labels are decoded
for up to four tokens.

Representation probes use SST-2 splits of
9,600/2,400/872 and TweetEval splits of
24,942/1,131/6,347. Steering and full ablation
use 600 balanced examples per dataset. Matched
ablation uses an inspected TweetEval sample
and a separate SST-2 training holdout,
each containing 600 examples.

Correction classifiers train on 14,186 balanced
source examples and are evaluated on 5,000
SST-2 or 6,347 TweetEval examples. Frozen
transfer uses 20,000 Amazon reviews.

\subsection{Prompt Sensitivity and Emotion Validation}
\label{subsec:baseline_results}

\begin{table}[htbp]
\centering\small
\caption{Sentiment classification accuracy (\%) under
four prompt variants (P1--P4). SST-2 reports
generated-label accuracy on pooled sampled training
and validation examples. TweetEval reports binary
positive-negative sentiment evaluation.}
\label{tab:baseline_results}
\begin{tabular}{@{}llrrrr@{}}
\toprule
Model & Dataset & P1 & P2 & P3 & P4 \\
\midrule
0.5B & SST-2     & 89.44 & 86.09 & 85.82 & 89.05 \\
3B   & SST-2     & 79.18 & 85.26 & 81.46 & 87.80 \\
0.5B & TweetEval & 79.62 & 78.92 & 77.31 & 82.72 \\
3B   & TweetEval & 78.35 & 79.52 & 78.01 & 81.57 \\
\bottomrule
\end{tabular}
\end{table}

Prompt wording affects generated-label accuracy
(Table~\ref{tab:baseline_results}). The saved
P1 instruction is used for downstream sentiment
correction and cross-domain transfer.
Prompt-sensitivity analysis evaluates generated
labels, whereas correction uses margin-based
binary classifiers. TweetEval evaluation
examples were inspected in earlier stages,
so the later correction results are treated
as exploratory.

Neutral-PCA denoising improves emotion
discrimination on the official
GoEmotions validation split. Across
all layers and eight one-vs-rest
tasks, mean AUROC rises from 0.616
to 0.883 for 0.5B. For 3B, it rises
from 0.611 to 0.891. This evaluation
uses 879 strict single-label examples.
At the selected layers, the leading neutral
principal component explains 99.74\% of
neutral activation variance in 0.5B and
99.12\% in 3B. Thus thresholds between
30\% and 70\% would retain that same
component at these fixed layers. This
observation does not test whether selecting
layers again at a different threshold would
give the same result.

For layer selection, a seven-emotion evaluation uses 712 GoEmotions validation examples, excluding love, together with 490 EmpatheticDialogues validation examples. At the selected layers, denoised macro-AUROC reaches 0.876 for Qwen2.5-0.5B at layer 14 and 0.896 for Qwen2.5-3B at layer 18 on the matched GoEmotions subset. Each emotion is evaluated against the other six. These results are descriptive because validation performance also guided layer selection.

\subsection{Sentiment Information and Representation Geometry}
\label{subsec:sentiment_decoding_results}
\label{subsec:geometry_results}

\
\begin{table}[htbp]
\centering\footnotesize
\setlength{\tabcolsep}{3pt}
\caption{Sentiment prediction from representation-only
probes. Accuracy is a percentage; AUROC ranges from
0 to 1. Random results average ten seeds.}
\label{tab:probe_results}
\begin{tabular}{@{}llrrrr@{}}
\toprule
 & & \multicolumn{2}{c}{SST-2}
 & \multicolumn{2}{c}{TweetEval} \\
\cmidrule(lr){3-4}
\cmidrule(l){5-6}
Model & Features & Acc. (\%) & AUROC
& Acc. (\%) & AUROC \\
\midrule
0.5B & Raw emotion   & 81.88 & 0.898 & 88.66 & 0.958 \\
     & Denoised      & 81.77 & 0.892 & 87.99 & 0.957 \\
     & Random rank-7 & 63.08 & 0.695 & 53.11 & 0.740 \\
     & Full state    & 89.11 & 0.959 & 91.70 & 0.974 \\
\midrule
3B   & Raw emotion   & 79.13 & 0.868 & 90.70 & 0.961 \\
     & Denoised      & 78.21 & 0.870 & 90.09 & 0.961 \\
     & Random rank-7 & 62.00 & 0.675 & 54.22 & 0.753 \\
     & Full state    & 91.28 & 0.969 & 93.46 & 0.982 \\
\bottomrule
\end{tabular}
\end{table}

These probes predict sentiment from
model representations alone. They do
not use the model's output margin.
On binary TweetEval, raw emotion
features reach 88.66\% accuracy
for 0.5B and 90.70\% for 3B
(Table~\ref{tab:probe_results}).
Both exceed the random controls.
The full hidden state gives higher
accuracy still.

On SST-2, raw emotion features
reach 81.88\% and 79.13\%
accuracy. Denoising improves
emotion validation, but it does
not consistently improve
sentiment probing.

\begin{table}[htbp]
\centering\scriptsize
\setlength{\tabcolsep}{2.5pt}
\caption{Relationship between emotion and sentiment
directions. Coverage (\%) measures the fraction
of the supervised sentiment direction captured
by each subspace. Random coverage reports
mean $\pm$ SD across 10 rank-seven random
subspaces. Cosine similarity measures valence
alignment. Bold compares raw and denoised results.}
\label{tab:geometry_results}
\begin{tabular}{@{}llccccc@{}}
\toprule
 & & \multicolumn{3}{c}{Coverage (\%)}
 & \multicolumn{2}{c}{Cosine similarity} \\
\cmidrule(lr){3-5}
\cmidrule(l){6-7}
Model & Dataset & Raw & Denoised & Random
      & Raw & Denoised \\
\midrule
0.5B & SST-2
 & 49.54 & \textbf{50.22} & $0.81\pm0.28$
 & 0.041 & \textbf{0.599} \\

3B & SST-2
 & \textbf{40.91} & 40.51 & $0.36\pm0.17$
 & 0.232 & \textbf{0.551} \\

0.5B & TweetEval
 & 59.36 & \textbf{66.85} & $1.09\pm0.35$
 & 0.076 & \textbf{0.717} \\

3B & TweetEval
 & 57.41 & \textbf{62.91} & $0.46\pm0.24$
 & 0.166 & \textbf{0.664} \\
\bottomrule
\end{tabular}
\end{table}

Emotion-space coverage ranges from 40.51\%
to 66.85\%, substantially exceeding the
empirical random rank-seven controls.
The theoretical isotropic expectations are
0.78\% for 0.5B and 0.34\% for 3B.
This substantial overlap with supervised
sentiment directions may partly explain
the observed intervention effects without
establishing a uniquely emotional mechanism.

Denoising increases the cosine
similarity between emotion-based
valence and the sentiment direction
(Table~\ref{tab:geometry_results}).
On TweetEval, similarity rises
from 0.076 to 0.717 for 0.5B
and from 0.166 to 0.664 for 3B.
The denoised emotion spaces
also cover 66.85\% and 62.91\%
of the corresponding sentiment
directions.

Geometry and probe accuracy answer
different questions. Closer alignment
does not always produce better
classification. A successful probe
shows that a classifier can recover
sentiment information. It does not
show whether unmodified Qwen uses
that information when predicting.

\subsection{Interventions on Sentiment Predictions}
\label{subsec:steering_results}
\label{subsec:full_ablation_results}
\label{subsec:matched_ablation_results}

Steering tests whether emotion-related activation
changes affect output sentiment margins. At the
selected layer, interventions use 1\%, 2.5\%,
and 5\% of each valid token's activation norm.
Table~\ref{tab:steering_results} reports two-sided
responses. Denoised valence exceeds the maximum
absolute response of 50 random directions at
every strength and setting, whereas raw valence
remains below that maximum. The supervised
sentiment direction consistently gives the
largest response.

\begin{table}[htbp]
\centering\scriptsize
\setlength{\tabcolsep}{3pt}
\caption{Two-sided sentiment-margin responses
at three steering strengths. Random statistics
summarize absolute mean responses across 50
directions. Values are margin changes, not
accuracy changes.}
\label{tab:steering_results}
\begin{tabular}{@{}llcrrrrr@{}}
\toprule
Model & Dataset & Strength & Raw & Denoised &
Sentiment & Rand. med. & Rand. max. \\
\midrule
0.5B & SST-2 & 1\% & 0.020 & 0.160 & 0.237 & 0.008 & 0.043 \\
     &       & 2.5\% & 0.051 & 0.400 & 0.593 & 0.021 & 0.110 \\
     &       & 5\% & 0.101 & 0.799 & 1.185 & 0.042 & 0.219 \\
\midrule
3B & SST-2 & 1\% & 0.068 & 0.230 & 0.421 & 0.028 & 0.095 \\
   &       & 2.5\% & 0.171 & 0.582 & 1.083 & 0.070 & 0.236 \\
   &       & 5\% & 0.342 & 1.208 & 2.326 & 0.140 & 0.473 \\
\midrule
0.5B & TweetEval & 1\% & 0.023 & 0.079 & 0.080 & 0.011 & 0.038 \\
     &           & 2.5\% & 0.057 & 0.198 & 0.199 & 0.029 & 0.094 \\
     &           & 5\% & 0.112 & 0.393 & 0.398 & 0.057 & 0.187 \\
\midrule
3B & TweetEval & 1\% & 0.056 & 0.240 & 0.259 & 0.027 & 0.118 \\
   &           & 2.5\% & 0.138 & 0.603 & 0.648 & 0.067 & 0.298 \\
   &           & 5\% & 0.276 & 1.214 & 1.305 & 0.136 & 0.590 \\
\bottomrule
\end{tabular}
\end{table}

Individual emotion directions also affect margins
in the expected direction. On TweetEval, denoised
joy produces responses of 0.370 and 1.072 for
0.5B and 3B, while sadness produces $-0.395$
and $-0.819$. All eight denoised emotions show
the expected response direction. These effects
demonstrate behavioral sensitivity, not that
the model normally relies on these directions.

Full emotion-space removal is evaluated using
binary candidate-token accuracy
(Table~\ref{tab:full_ablation_results}).
Its separate evaluation samples and scoring
protocol prevent direct comparison with
prompt-sensitivity or correction accuracies.

\begin{table}[htbp]
\centering\small
\caption{Accuracy (\%) after full emotion-space
removal. Parentheses indicate percentage-point
drops from the unmodified baseline.}
\label{tab:full_ablation_results}
\begin{tabular}{@{}llccc@{}}
\toprule
Model & Dataset & Baseline & Raw & Denoised \\
\midrule
0.5B & SST-2     & 89.50 & 88.00 (1.50) & 88.67 (0.83) \\
3B   & SST-2     & 91.00 & 86.83 (4.17) & 90.67 (0.33) \\
0.5B & TweetEval & 87.17 & 77.83 (9.33) & 78.50 (8.67) \\
3B   & TweetEval & 87.50 & 77.83 (9.67) & 80.50 (7.00) \\
\bottomrule
\end{tabular}
\end{table}

The SST-2 full- and matched-ablation experiments
use distinct, non-overlapping balanced samples
of 600 examples. Full ablation uses official
validation data (3B baseline: 91.00\%), whereas
matched ablation uses an internal training holdout
(81.67\%). Both retain the same two-label P1
prompt and binary candidate-token scoring.
The discrepancy persists across larger cached
prediction pools and is not specific to the
600-example samples. These results cannot be
interpreted as a same-sample comparison.

Full raw removal causes the largest accuracy
drops on TweetEval: 9.33 and 9.67 percentage
points for 0.5B and 3B, respectively. On SST-2,
the 3B raw and denoised drops are 4.17 and
0.33 points.

By comparison, full random rank-seven removal
reduces TweetEval accuracy by 0.78 and 0.46
points. However, equal-rank removal does not
imply equal perturbation energy. For 0.5B,
raw, denoised, and random removal change
4.66\%, 6.70\%, and 0.77\% of normalized
activation energy; the corresponding 3B values
are 8.10\%, 3.93\%, and 0.35\%. Random
energies approximate the isotropic expectations
$7/896$ and $7/2048$. Thus, equal-rank
comparisons alone cannot establish superiority
over other data-derived subspaces.

Energy-matched ablation addresses this difference
by scaling emotion-space removal separately for
each example to match random-removal energy.
TweetEval results use binary candidate scores
reconstructed from saved intervention outputs,
without changing model inputs or interventions.
Results average 50 paired random seeds rather
than treating repeated predictions as independent.

\begin{table}[htbp]
\centering\small
\caption{TweetEval matched ablation with binary
candidate scoring. Excess loss is the additional
accuracy drop versus random removal (percentage
points). Results average 50 paired seeds. 
Excess losses are computed before rounding,
 brackets give nominal paired 95\% percentile CIs
from 50,000 bootstrap resamples.}
\label{tab:matched_results}
\begin{tabular}{@{}llrl@{}}
\toprule
Model & Removal & Acc. (\%) & Excess loss [95\% CI] \\
\midrule
0.5B & None     & 87.17 & -- \\
     & Random   & 86.39 & -- \\
     & Raw      & 84.00 & \textbf{2.39} [1.21, 3.67] \\
     & Denoised & 84.16 & \textbf{2.23} [1.21, 3.35] \\
\midrule
3B   & None     & 87.50 & -- \\
     & Random   & 87.04 & -- \\
     & Raw      & 85.94 & \textbf{1.10} [0.22, 2.05] \\
     & Denoised & 85.84 & \textbf{1.19} [0.24, 2.23] \\
\bottomrule
\end{tabular}
\end{table}

\begin{figure}[!htbp]
\centering
\includegraphics[width=0.96\linewidth]
{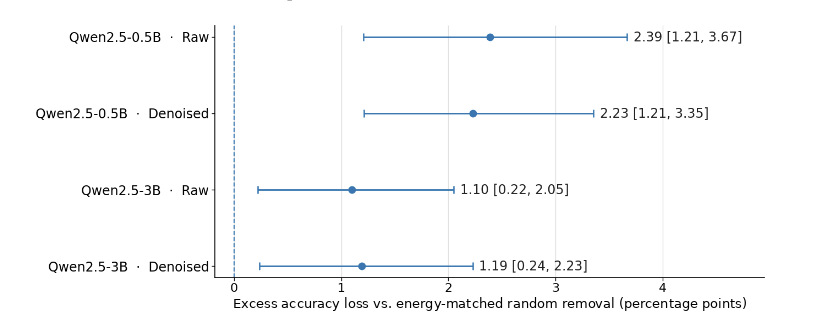}
\caption{Excess accuracy loss relative to
energy-matched random removal. Points show
percentage-point effects; bars show nominal
paired 95\% percentile CIs from 50,000 resamples.
The dashed line indicates zero.}
\label{fig:distortion_matched}
\end{figure}

At matched energy, raw-emotion removal produces
2.39 and 1.10 points of excess loss for 0.5B
and 3B (Table~\ref{tab:matched_results}).
Both nominal intervals exclude zero.
Figure~\ref{fig:distortion_matched} visualizes
all four effects; family-wise inference appears
in Table~\ref{tab:adjusted_ablation}.

These effects apply to the tested isotropic
random controls, not all data-derived subspaces.
Emotion spaces substantially overlap supervised
sentiment directions, potentially explaining
some intervention effects without a uniquely
emotional mechanism. Energy matching controls
perturbation magnitude, not direction or
downstream consequences.

The exploratory SST-2 extension uses 600
balanced official-training examples with no
audited source-ID or normalized-text overlap
against previous saved experimental uses.
It retains the original P1 prompt and binary
scoring, with 50 paired random controls.
The initial 2,000-draw analysis was reproduced,
and intervals use 50,000 class-stratified paired
example-and-seed bootstrap resamples.
All 30,000 energy comparisons passed for each
model and emotion-space condition. Random-removal
accuracy SDs were 0.29 and 0.57 points for
0.5B and 3B, respectively.

On SST-2, raw removal yields excess losses
of 0.66 and 0.57 points. Only the 3B nominal
raw interval excludes zero; neither survives
Holm adjustment. These effects are smaller
than TweetEval's, but differing evaluation
samples prevent attributing the difference
to dataset properties alone. Both analyses
are exploratory.

\begin{table}[htbp]
\centering\small
\setlength{\tabcolsep}{4pt}
\caption{Exploratory SST-2 matched ablation
on 600 balanced training-holdout examples.
Accuracy averages 50 seeds; excess loss is
relative to random removal (percentage points).
Brackets show nominal paired 95\% CIs.}
\label{tab:sst2_matched}
\begin{tabular}{@{}llrl@{}}
\toprule
Model & Removal & Acc. (\%) & Excess loss [95\% CI] \\
\midrule
0.5B & None & 90.00 & -- \\
     & Random & 89.81 & -- \\
     & Raw & 89.15 & 0.66 [-0.10, 1.49] \\
     & Denoised & 89.66 & 0.15 [-0.22, 0.61] \\
\midrule
3B   & None & 81.67 & -- \\
     & Random & 81.70 & -- \\
     & Raw & 81.14 & 0.57 [0.12, 1.09] \\
     & Denoised & 81.30 & 0.40 [-0.32, 1.15] \\
\bottomrule
\end{tabular}
\end{table}

Multiple-comparison-adjusted results for
both datasets appear in
Table~\ref{tab:adjusted_ablation}.

\begin{table}[htbp]
\centering\scriptsize
\setlength{\tabcolsep}{3pt}
\caption{Matched-ablation inference across
eight comparisons. Bonferroni basic-bootstrap
intervals target approximate 95\% family-wise
coverage; Holm-adjusted $p$-values use 50,000
paired resamples. Intervals are not
Holm-adjusted. Effects are percentage points.}
\label{tab:adjusted_ablation}
\begin{tabular}{@{}lllcr@{}}
\toprule
Dataset & Model & Removal & Family-wise 95\% CI & Holm $p$ \\
\midrule
TweetEval & 0.5B & Raw & [0.57, 4.02] & \textbf{0.0021} \\
          &      & Denoised & [0.60, 3.60] & \textbf{0.0008} \\
          & 3B   & Raw & [-0.25, 2.31] & \textbf{0.042} \\
          &      & Denoised & [-0.27, 2.53] & \textbf{0.048} \\
\midrule
SST-2     & 0.5B & Raw & [-0.53, 1.72] & 0.2982 \\
          &      & Denoised & [-0.54, 0.65] & 0.5661 \\
          & 3B   & Raw & [-0.20, 1.19] & 0.1102 \\
          &      & Denoised & [-0.69, 1.40] & 0.5661 \\
\bottomrule
\end{tabular}
\end{table}

All four exploratory TweetEval effects have
Holm-adjusted $p<0.05$ (0.0021, 0.0008,
0.042, and 0.048), while none of the four
SST-2 effects survives adjustment.
Bonferroni simultaneous intervals exclude
zero for the two 0.5B TweetEval effects,
but not for 3B. These interval and test
procedures therefore give different
conclusions for 3B. The approximate bootstrap
$p$-values have finite Monte Carlo precision.
Overall, the evidence indicates sensitivity
to tested isotropic random-space comparisons
on TweetEval, not a uniquely emotional
mechanism or robust cross-dataset generalization.

\subsection{Sentiment Correction}
\label{subsec:correction_results}

To assess predictive utility beyond Qwen's output
margin, source-trained logistic-regression
classifiers combine margin and emotion features.
Raw, denoised, and random features are compared
with margin-only correction and uncorrected Qwen.
Each setting retains its saved P1 prompt;
preprocessing and model selection use source
training and validation data only.

\begin{table}[htbp]
\centering\small
\setlength{\tabcolsep}{3pt}
\caption{Source-domain accuracy (\%).
Random results average 50 classifiers;
the final row reports across-seed SD
in percentage points. Bold marks the best.}
\label{tab:stagei}
\begin{tabular}{@{}lrrrr@{}}
\toprule
 & \multicolumn{2}{c}{SST-2} &
   \multicolumn{2}{c}{TweetEval} \\
\cmidrule(lr){2-3}
\cmidrule(l){4-5}
Method & 0.5B & 3B & 0.5B & 3B \\
\midrule
Original Qwen        & 89.32 & 80.96 & 85.88 & 89.81 \\
Margin only          & 89.14 & 91.34 & 85.88 & 92.48 \\
Raw emotion + margin & \textbf{90.18} & \textbf{91.44}
                     & \textbf{92.14} & \textbf{94.03} \\
Denoised + margin    & 89.88 & 91.28 & 92.09 & 94.01 \\
Random + margin      & 89.75 & 91.24 & 86.93 & 92.50 \\
\midrule
Random seed SD       & 0.23 & 0.17 & 0.87 & 0.19 \\
\bottomrule
\end{tabular}
\end{table}

Raw emotion features outperform margin-only
correction most strongly on TweetEval
(Table~\ref{tab:stagei}), improving accuracy
by 6.26 and 1.55 percentage points for
0.5B and 3B, respectively (paired 95\% CIs:
[5.47, 7.07] and [1.10, 2.02]).
Both also outperform random-feature controls,
with AUROCs of 0.9706 and 0.9836.

On SST-2, the corresponding gains are
1.04 points (95\% CI [0.56, 1.52])
and 0.10 points ([-0.36, 0.58]);
the latter is inconclusive.
These improvements establish predictive
utility for trained classifiers, not causal
reliance by unmodified Qwen.

\subsection{Frozen Cross-Domain Transfer}
\label{subsec:amazon_results}

Source-trained classifiers are evaluated
without target retraining or tuning on
20,000 balanced Amazon Polarity test reviews.
Prompts, layers, feature transforms, classifier
weights, and decision rules remain frozen.
Target labels determine sample balance only.

\begin{table}[htbp]
\centering\scriptsize
\setlength{\tabcolsep}{3pt}
\caption{Amazon accuracy/macro-$F_1$ (\%).
Random results average 50 source-trained
classifiers; the final row reports across-seed
SD in percentage points. Bold marks the best.}
\label{tab:amazon}
\begin{tabular}{@{}lcccc@{}}
\toprule
 & \multicolumn{2}{c}{SST-2 trained}
 & \multicolumn{2}{c}{TweetEval trained} \\
\cmidrule(lr){2-3}
\cmidrule(l){4-5}
Method & 0.5B & 3B & 0.5B & 3B \\
\midrule
Original Qwen
 & 92.24/92.22 & 94.13/94.12
 & 92.13/92.13 & 94.67/94.66 \\

Margin only
 & 92.05/92.02 & 95.86/95.86
 & 92.13/92.13 & \textbf{95.78/95.78} \\

Raw emotion + margin
 & \textbf{93.63/93.62} & \textbf{95.90/95.90}
 & 92.65/92.64 & 95.48/95.48 \\

Denoised + margin
 & 93.42/93.41 & 95.86/95.86
 & \textbf{92.93/92.93} & 95.39/95.38 \\

Random + margin
 & 92.57/92.56 & 95.85/95.86
 & 92.24/92.23 & 95.68/95.68 \\
\midrule
Random seed SD
 & 0.29/0.30 & 0.05/0.05
 & 1.35/1.37 & 0.12/0.12 \\
\bottomrule
\end{tabular}
\end{table}

\begin{table}[htbp]
\centering\small
\setlength{\tabcolsep}{4pt}
\caption{Amazon raw-emotion accuracy gains
(percentage points), with nominal paired
95\% percentile CIs from 2,000 resamples.}
\label{tab:amazon_ci}
\begin{tabular}{@{}llcc@{}}
\toprule
Source & Model & Raw -- Margin & Raw -- Original \\
\midrule
SST-2 & 0.5B
 & +1.58 [1.34, 1.83]
 & +1.39 [1.15, 1.62] \\
SST-2 & 3B
 & +0.04 [-0.08, 0.17]
 & +1.77 [1.50, 2.04] \\
TweetEval & 0.5B
 & +0.52 [0.22, 0.82]
 & +0.52 [0.22, 0.82] \\
TweetEval & 3B
 & -0.30 [-0.47, -0.13]
 & +0.81 [0.57, 1.03] \\
\bottomrule
\end{tabular}
\end{table}

Raw-emotion correction improves over
uncorrected Qwen in all four Amazon settings
by 0.52--1.77 points
(Tables~\ref{tab:amazon} and
\ref{tab:amazon_ci}). However, relative to
margin-only correction, gains are inconsistent:
+1.58 and +0.04 points for SST-2-trained
0.5B and 3B, and +0.52 and -0.30 points
for TweetEval-trained models. The +0.04-point
gain is inconclusive, while the negative
TweetEval 3B difference indicates worse accuracy.

Macro-$F_1$ shows broadly similar rankings.
Accuracy gains also need not improve AUROC:
SST-2-trained 3B raw correction achieves
0.9874, below uncorrected Qwen's 0.9888.
Thus, emotion features offer some cross-domain
benefit but no consistent advantage over
margin-only correction. Source-specific
results are not directly interchangeable.

\subsection{Discussion}
\label{subsec:results_discussion}

Emotion probes recover substantial sentiment
information, but improved geometric alignment
after neutral-PCA denoising does not consistently
improve classification. Probe accuracy alone
does not establish reliance by the original model.

Steering and ablation affect model outputs.
Raw valence steering remains below the maximum
tested random response, whereas denoised valence
exceeds it. All four exploratory TweetEval
matched-ablation effects survive Holm correction,
but only the two 0.5B effects have simultaneous
Bonferroni-adjusted intervals excluding zero.
No SST-2 effect survives Holm correction.
These findings show sensitivity relative to
isotropic random controls, not a uniquely
emotional mechanism.

Emotion-based correction improves some predictions
but does not consistently outperform margin-only
correction during frozen Amazon transfer.

For trustworthy AI, these results highlight
the limits of representation-based explanations.
Future work should examine additional model
families, independent datasets, data-aligned
controls, and the relationship between
denoising and predictive performance.

\subsection{Limitations}
\label{subsec:limitations}

The experiments examine only two models from the
same model family and use selected internal layers.
The historical TweetEval matched-ablation sample
was inspected before the final comparison. The
SST-2 extension has no audited overlap with
prior saved examples, but broader independence
from undocumented decisions cannot be established.
Its results are therefore exploratory.
Individual 95\% percentile confidence intervals
are nominal, whereas Bonferroni-adjusted
basic-bootstrap simultaneous intervals target
approximate 95\% family-wise coverage across
eight comparisons. These intervals are distinct
from Holm-adjusted $p$-values. Matched perturbation energy
does not control intervention direction or all
possible distribution shifts in the activations.

The generated-label prompts, binary candidate-token intervention readout, and binary correction heads use different prediction rules.
Prompt-sensitivity results are descriptive
and are not directly comparable with the
correction experiments. Neutral-PCA uses
a fixed explained-variance threshold, and
a full threshold-sensitivity analysis was
not conducted. These limitations restrict
the interpretation and generalization
of the findings.

\section{Conclusion}
\label{sec:conclusion}
Emotion directions in the two Qwen2.5 models
contain sentiment information recoverable by
supervised probes. Steering and ablation show
that changing these directions can affect model
outputs. In matched ablation, all four TweetEval
effects survive Holm correction across eight
comparisons, whereas none of the SST-2 effects
does. Only the two TweetEval 0.5B effects have
Bonferroni-adjusted simultaneous intervals
excluding zero. Neutral-PCA denoising
improves emotion validation and valence alignment
without consistently improving sentiment prediction.
Frozen emotion-based correction improves over
uncorrected Qwen on Amazon, but does not consistently
outperform margin-only correction. The results
therefore distinguish accessible sentiment
information, intervention effects, and practical
predictive utility. They do not establish a
uniquely emotional mechanism or broad transfer
to other model families.

\section*{Data Rights, Ethics and Model Licenses}

This study uses five publicly available datasets:
GoEmotions (Apache 2.0), SST-2 (license unspecified),
TweetEval (license unspecified, subject to platform
terms), EmpatheticDialogues (CC BY-NC 4.0), and
Amazon Polarity (Hugging Face release listed as
Apache 2.0). Dataset licenses may not cover rights
in underlying user-generated content. The models
are Qwen2.5-0.5B-Instruct (Apache 2.0) and
Qwen2.5-3B-Instruct (Qwen Research License).
The source code is publicly available at
\url{https://github.com/Alihassan0155/Do-Language-Models-Feel-the-Sentiment}.
Datasets, individual predictions, evaluation
identifiers, extracted emotion vectors, trained
correction heads, and experimental artifacts are
not redistributed. Researchers should obtain
datasets and model checkpoints from their original
providers under the applicable terms.

\section*{Declaration on Generative AI}

During the preparation of this work, the authors used
ChatGPT (OpenAI) for language editing, academic
rephrasing, and suggestions regarding figure layouts.
All AI-assisted outputs were reviewed and edited by
the authors, who take full responsibility for the
scientific content, results, interpretations, and
conclusions. Figures 1 and 2 were manually designed by the authors
using Canva, with no generative AI-generated artwork.
Figure 3 was plotted directly from the experimental
results without the use of generative AI tools.

\bibliography{references}
\end{document}